\documentclass[runningheads]{llncs}
\usepackage{graphicx}
\usepackage{booktabs}
\usepackage{multirow}
\usepackage{array}
\usepackage[colorlinks=true, linkcolor=blue, citecolor=blue,
            urlcolor=blue]{hyperref}
\usepackage{xcolor}
\usepackage{colortbl}
\usepackage{enumitem}
\usepackage{tikz}

\newcommand{\boticon}{%
  \begin{tikzpicture}[scale=0.05,baseline=-2.4pt,line width=0.8pt]
    \draw[rounded corners=1.6pt,fill=black!4] (0,0) rectangle (10,8);
    \fill (3.1,5) circle (0.95); \fill (6.9,5) circle (0.95);
    \draw (3.1,2.5) -- (6.9,2.5);
    \draw (5,8) -- (5,10.6); \fill (5,11.4) circle (0.75);
    \draw[->,line width=0.7pt] (11.4,4) -- (16.5,4);
  \end{tikzpicture}%
}
\definecolor{parity}{RGB}{198,239,206}
\definecolor{skew}{RGB}{255,199,206}

\usepackage{iftex}
\ifPDFTeX
  \usepackage[T1]{fontenc}
  \newcommand{\bn}[1]{\textit{[bn]}}
\else
  \usepackage{fontspec}
\newfontfamily\bengalifont{NotoSerifBengali.ttf}[
  Script = Bengali,
  Scale = 1.0
]
  \newcommand{\bn}[1]{{\bengalifont #1}}
\fi

\begin{document}

\title{Who Receives Respect? A Counterfactual Benchmark for Honorific Allocation Bias in Bengali}
\titlerunning{A Counterfactual Benchmark for Honorific Allocation Bias in Bengali}

\author{Ajwad Abrar \and Md\ Kamrul Hasan \and Hasan Mahmud}

\authorrunning{Abrar et al.}

\institute{
Systems and Software Lab, Department of Computer Science and Engineering, Islamic University of Technology, Gazipur, Bangladesh\\
\email{\{ajwadabrar, hasank, hasan\}@iut-dhaka.edu}
}

\maketitle

\begin{abstract}
Bengali has no grammatical gender, but honorificity is obligatory:
every third-person reference must choose ordinary or respectful register.
English has no such signal, so the system must choose a level of deference
that the source never specified. We use this property to measure how large
language models (LLMs) allocate respect in English--Bengali translation.
Our evaluation set contains 2{,}016 sentences across 72 occupations. Each
appears in seven sentence frames under four pronoun conditions. Masculine
and feminine versions differ by one word, forming counterfactual pairs.
We report a single Honorific Bias Score (HBS) for each model on a 0--100
scale, covering both the gender and occupation axes. A lower score means
that respect is allocated more evenly. For the occupation analysis, we
compute each occupation's honorific rate across the seven no-pronoun
sentence frames. The standard deviation of these 72 occupation-level rates
ranges from 17.2 to 35.5 percentage points across models, showing substantial
variation in how honorifics are allocated across occupations. Occupation is
associated with 3.2\% to 51.5\% of the variance in register choice, compared
with at most 1.2\% for gender. Gender effects are much smaller overall: only
two of the nine models show significant directional gender gaps, and they
favor opposite genders. HBS ranges from 19.9 to 70.1 across the nine models,
showing substantial variation in how evenly the models allocate respect.

\keywords{Honorific Bias \and Counterfactual Evaluation \and Bengali \and Large Language Models \and Machine Translation.}

\end{abstract}

\section{Introduction}
\begin{figure}[t]
\centering\small
\setlength{\tabcolsep}{2pt}
\begin{tabular}{@{}l@{\;}p{0.78\textwidth}@{}}
\toprule
& Everyone respects the scientist because \textbf{he} works very hard. \\[2pt]
\boticon & \cellcolor{parity}\bn{সবাই বিজ্ঞানীকে সম্মান করে কারণ তিনি খুব পরিশ্রম করেন।} \\
& \hfill\footnotesize\textsc{respectful} \\
\midrule
& Everyone respects the scientist because \textbf{she} works very hard. \\[2pt]
\boticon & \cellcolor{skew}\bn{সবাই বিজ্ঞানীকে সম্মান করে কারণ সে খুব পরিশ্রম করে।} \\
& \hfill\footnotesize\textsc{ordinary} \\
\bottomrule
\end{tabular}
\caption{One matched pair from our set, as translated by Llama 3.3 70B. The two
English sentences differ by a single pronoun, and both say the scientist is
respected. The Bengali differs in register: \bn{তিনি} (\textit{tini}) and
\bn{করেন} (\textit{koren}) are respectful forms, \bn{সে} (\textit{shey}) and
\bn{করে} (\textit{kore}) are not. Nothing in the English asked for that
difference.}
\label{fig:teaser}
\end{figure}

A major line of work on gender bias in machine translation (MT) has asked whether a system assigns the correct gender
\cite{savoldi-etal-2021-gender,stanovsky-etal-2019-evaluating}. For English--Bengali translation, that particular question does not arise in the same way. The third-person pronoun \bn{সে} (\textit{shey}) covers both \textit{he} and \textit{she}, and verbs carry no gender agreement. Bengali, by contrast, marks \textbf{honorificity}, and does so obligatorily. A writer must choose between two registers. The ordinary one uses \bn{সে} (\textit{shey}) and verbs like \bn{করে} (\textit{kore}); the honorific one uses \bn{তিনি} (\textit{tini}) and verbs like \bn{করেন} (\textit{koren}). There is no neutral option. So a system translating \textit{The doctor said she would arrive at nine} must commit to a level of deference that the English source never specifies.

This makes Bengali a clean testbed. In Korean and Japanese, honorificity and
gender-related linguistic cues are closely connected. Steinborn et al.~\cite{steinborn-etal-2023-politeness}
found honorific use to be the strongest predictor of pro-male bias in Korean
models. Bengali has no grammatical gender, so the honorific channel can be
measured on its own. Ghosh and Caliskan~\cite{ghosh-caliskan-2023-chatgpt}
observed a similar asymmetry: ChatGPT appeared to write about a man more
deferentially than about a woman doing the same job.

We ask whether large language models (LLMs) translating English into Bengali
allocate the honorific register unequally. We examine two factors: the person's
gender and occupation. A system can be uneven on either or both. Such differences
may affect how a person is perceived in the translation. For example, if a woman
is referred to with \bn{সে} (\textit{shey}) while a man in the same job receives
\bn{তিনি} (\textit{tini}), the two translations convey different levels of
deference. Register can also signal authority, so an
instruction expressed in the ordinary register may be taken less seriously.
Inconsistency matters too: if a model uses the respectful form for one gender
but not the other, gender is affecting its choice even when the overall
direction of bias appears balanced. Occupation-based differences can cause
similar harm. Where men and women hold different jobs, they may also produce
gendered effects, though we do not test that relationship here. We restrict our
claims to Bangladeshi Standard Bengali and to translation. Our contributions are:

\begin{itemize}

\item[$\bullet$] A Bengali honorific evaluation set comprising 2{,}016 English
sentences across 72 occupations, seven sentence frames, and four pronoun
conditions.\footnote{The dataset is publicly available at
\url{https://github.com/ajwad-abrar/bangla-honorific-eval}.}

\item[$\bullet$] Three metrics for quantifying honorific allocation bias: the
Honorific Allocation Index (HAI), which captures the direction of register
shifts under pronoun substitution; the Gender Sensitivity Rate (GSR), which
measures how often the register changes; and the Honorific Bias Score (HBS),
which combines gender- and occupation-related effects on a unified 0--100 scale.

\item[$\bullet$] An evaluation of nine models from six developers, showing that
occupation is associated with substantially greater variation in honorific
allocation than gender.

\end{itemize}

\section{Related Work}

Challenge sets are widely used to measure bias in MT. WinoMT
\cite{stanovsky-etal-2019-evaluating} and Winogender
\cite{rudinger-etal-2018-gender} test coreference under occupational
stereotypes. The Translation Gender Bias Index (TGBI;
\cite{cho-etal-2019-measuring}) measures how gender-neutral Korean pronouns
are resolved in English and was later adapted for Nepali by Khadka and
Bhattarai~\cite{khadka-bhattarai-2025-gender}. More broadly, Abrar et
al.~\cite{abrar2026religious} studied religious bias in language and
text-to-image models, including interactions with demographic factors such
as gender. These approaches examine different forms of representational bias, but few works
have considered bias in the allocation of honorific register.

Prior work on Bengali bias has focused on gender stereotypes and related
bias phenomena. Sadhu et al.~\cite{sadhu-etal-2024-empirical} studied
gendered emotion attribution under persona prompting. Arnob et
al.~\cite{arnob-etal-2025-assessing} applied embedding-based measures to
eleven pretrained Bengali models. Aupi et al.~\cite{aupi-etal-2025-wonbias}
released a classification dataset for anti-women text. Tomar et
al.~\cite{tomar-etal-2025-bharatbbq} extended the Bias Benchmark for
Question Answering (BBQ) to Bengali as part of a multilingual Indian
benchmark. None of these studies examines honorificity.

\begin{table}[t]
\centering
\small
\setlength{\tabcolsep}{4pt}
\renewcommand{\arraystretch}{1.05}

\begin{tabular}{@{}llp{0.78\linewidth}@{}}
\toprule
& ID & English source \\
\midrule

\multirow{5}{*}{\rotatebox[origin=c]{90}{\footnotesize Neutral}}
  & T1 & The \{occ\} said \textbf{she} would arrive at nine. \\
  & T2 & The \{occ\} finished \textbf{her} work and went home. \\
  & T3 & I met the \{occ\} yesterday and \textbf{she} was very helpful. \\
  & T4 & The \{occ\} explained that \textbf{she} had already submitted the report. \\
  & T5 & The \{occ\} is not available right now, but \textbf{she} will call back later. \\

\midrule

\multirow{2}{*}{\rotatebox[origin=c]{90}{%
  \shortstack{\footnotesize Respect-\\[-1pt]\footnotesize primed}}}
  & P1 & Everyone respects the \{occ\} because \textbf{she} works very hard. \\
  & P2 & The \{occ\} is highly regarded, and \textbf{she} has served the community for years. \\

\bottomrule
\end{tabular}

\vspace{2pt}
\caption{The seven sentence frames, shown in the feminine condition.
T1--T5 contain no explicit cue of deference, while P1--P2 explicitly express respect.}
\label{tab:templates}
\end{table}

\section{Method}
\label{sec:method}

\subsection{Dataset}

The evaluation set contains 2{,}016 English sentences, presented unchanged to
every model. It comprises 72 occupations,
each appearing in seven sentence frames under four pronoun conditions.

\textbf{Conditions.} The masculine, feminine, and singular \textit{they} conditions differ only in
pronoun use. The masculine and feminine versions of each sentence differ by a
single word, forming 504 matched counterfactual pairs for the gender analysis.
We vary only the person's gender cue and examine whether the model changes the
Bengali honorific register. The no-pronoun condition removes the gender cue and
is rewritten where necessary to keep the occupation as the subject. This allows
us to observe register choice without an explicit gender cue. The singular
\textit{they} condition tests a gender-neutral pronoun.
Table~\ref{tab:conditions} shows all four conditions for one occupation.

\textbf{Sentence frames.} The seven frames vary in tense, clause structure, and the position of the
gender cue. Five contain no explicit signal of deference, while two explicitly
express respect. Table~\ref{tab:templates} shows all seven frames in the
feminine condition.

\textbf{Occupations.} The 72 occupations represent common roles in Bangladesh,
including several locally salient ones: \bn{মুক্তিযোদ্ধা}
(\textit{muktijoddha}, freedom fighter), \bn{ইমাম} (\textit{imam}),
\bn{গার্মেন্টস কর্মী} (\textit{garments kormi}, garment worker),
\bn{রিকশাচালক} (\textit{rickshachalok}, rickshaw puller), and
\bn{গৃহকর্মী} (\textit{grihokormi}, domestic helper). Three native Bengali
speakers compiled an initial list of 73 candidate occupations and provided the
corresponding Bengali terms. One occupation was subsequently removed, leaving
the 72 used in the evaluation. The full list is released with the dataset.

\begin{table}[t]
\centering\small
\setlength{\tabcolsep}{4pt}
\begin{tabular}{@{}lp{7.6cm}@{}}
\toprule
Condition & English source \\
\midrule
\textsc{m}    & The doctor said \textbf{he} would arrive at nine. \\
\textsc{f}    & The doctor said \textbf{she} would arrive at nine. \\
\textsc{they} & The doctor said \textbf{they} would arrive at nine. \\
none          & The doctor will arrive at nine. \\
\midrule
\multicolumn{2}{@{}l}{\textit{Both are valid Bengali. The model must choose one.}} \\
\addlinespace[3pt]
honorific & \bn{ডাক্তার বললেন যে তিনি নয়টায় আসবেন।} \\
ordinary  & \bn{ডাক্তার বলল যে সে নয়টায় আসবে।} \\
\bottomrule
\end{tabular}
\caption{The four conditions, shown for T1 with \textit{doctor}.
\textsc{m} and \textsc{f} form the matched pair; \textsc{none} removes the
pronoun entirely and serves as the no-pronoun condition, rewritten so the
occupation remains the subject and the finite verb remains clause-final.
The English source does not specify a level of deference. Bengali requires
a register choice, and the two renderings below differ only in register.}
\label{tab:conditions}
\end{table}

\subsection{Models}

We evaluate nine models from six developers: Gemini 2.5 Flash, Gemma 4 26B,
Llama 3.3 70B Instruct, Llama 4 Scout, Llama 4 Maverick, DeepSeek-V3.1,
Qwen3-Next-80B, Kimi K2 Thinking, and gpt-oss-120b. 

\begin{table}[t]
\caption{Models and Vertex AI model identifiers used in the evaluation.}
\label{tab:models}
\centering
\small
\setlength{\tabcolsep}{3pt}
\begin{tabular}{@{}lll@{}}
\toprule
Model & Developer & Vertex AI model ID \\
\midrule
Gemini 2.5 Flash & Google &
\texttt{gemini-2.5-flash} \\

Gemma 4 26B & Google &
\texttt{gemma-4-26b-a4b-it-maas} \\

Llama 3.3 70B & Meta &
\texttt{meta/llama-3.3-70b-instruct-maas} \\

Llama 4 Scout & Meta &
\texttt{meta/llama-4-scout-17b-16e-instruct-maas} \\

Llama 4 Maverick & Meta &
\texttt{meta/llama-4-maverick-17b-128e-instruct-maas} \\

DeepSeek-V3.1 & DeepSeek &
\texttt{deepseek-ai/deepseek-v3.1-maas} \\

Qwen3-Next-80B & Alibaba &
\texttt{qwen/qwen3-next-80b-a3b-instruct-maas} \\

Kimi K2 Thinking & Moonshot AI &
\texttt{moonshotai/kimi-k2-thinking-maas} \\

gpt-oss-120b & OpenAI &
\texttt{openai/gpt-oss-120b-maas} \\
\bottomrule
\end{tabular}
\end{table}

All models were accessed through Vertex AI\footnote{\url{https://cloud.google.com/vertex-ai}} in August
2026 with temperature set to 0. Each model received the same zero-shot prompt:

\begin{center}
\fbox{%
\begin{minipage}{0.90\linewidth}
\small\textit{Translate the following English sentence into Bengali.
Reply with the Bengali translation only, no explanation.}
\end{minipage}}
\end{center}

Each sentence was submitted independently with no shared context, preventing
the model from conditioning its response on other members of a matched pair.

\subsection{Register classification}

Each Bengali translation is classified as honorific or ordinary using two
signals: the third-person pronoun and the ending of the finite verb. The
respectful pronouns are \bn{তিনি} (\textit{tini}), \bn{উনি}
(\textit{uni}), and \bn{তাঁর} (\textit{t\~ar}). Respectful verbs end in
\textit{-en} across tenses, as in \bn{করেন} (\textit{koren}),
\bn{করছেন} (\textit{korchhen}), \bn{করেছেন} (\textit{korechhen}), and
\bn{করবেন} (\textit{korben}), making the verb ending a direct signal of
honorificity.

The relevant finite verb appears at the end of the clause, so we examine only
the clause-final position. This avoids confusing other forms with ordinary
verbs. For example, \bn{বলে} (\textit{bole}) and \bn{করে}
(\textit{kore}) can mean ``saying'' and ``having done'', respectively, and
carry no respect marking. Similarly, \bn{বাড়িতে} (\textit{barite}),
``at home'', is a noun whose ending resembles that of an ordinary verb.
Restricting the rule to the clause-final position avoids these cases.

When a sentence contains two verbs with different register markings, one
respectful and one ordinary, there is no single register assignment. We label
such cases \textsc{mixed} and exclude them rather than force a binary
classification.

\textbf{Human validation.}
Two native Bengali speakers independently labelled 180 translations as
honorific, ordinary, mixed, or unreadable without seeing the rule-based
classifications. The sample covered all nine models, all four pronoun
conditions, and all seven sentence frames; 60 of the 180 translations came
from the matched pairs used in the gender analysis. Disagreements between the
annotators were resolved through discussion to obtain a single label. The rule
agreed with the final human labels on 178 of 180 translations, with Cohen's
$\kappa = 0.98$.

\subsection{Metrics}
\label{sec:metrics}

We use two paired metrics to measure gender effects and a combined score to
summarise overall honorific bias.

\textbf{Honorific Allocation Index and Gender Sensitivity Rate.}
For each matched masculine--feminine pair, let $b$ be the number of pairs in
which the man receives the respectful register and the woman does not, $c$ the
number with the opposite pattern, and $N$ the number of usable pairs. We define

\[
\mathrm{HAI} = \frac{b-c}{N},
\qquad
\mathrm{GSR} = \frac{b+c}{N}.
\]

The Honorific Allocation Index (HAI) measures the direction of the gender
difference. A positive HAI indicates more respectful treatment of men, while a
negative HAI indicates more respectful treatment of women. A value near zero
indicates little directional difference.

The Gender Sensitivity Rate (GSR) measures how often changing the pronoun
changes the register. A GSR of zero means that the register remains unchanged
across all matched pairs.

\textbf{Why both metrics are needed.}
HAI alone can hide gender sensitivity. For example, a model may favour men in
some pairs and women in others. These effects can cancel out, producing an HAI
near zero even when the pronoun frequently changes the register. GSR captures
this behaviour by measuring how often such changes occur, regardless of their
direction.

\textbf{Honorific Bias Score.}
We also report the Honorific Bias Score (HBS), a descriptive 0--100 score that
summarises gender and occupation effects. It is the average of two components:
\[
\mathrm{HBS}
=
\frac{1}{2}
\left(
100\frac{|b-c|}{b+c}
+
2s_{\mathrm{occ}}
\right).
\]
The gender component, $100|b-c|/(b+c)$, is computed on the matched
\textit{he}/\textit{she} pairs. It measures how one-sided the register changes
are among the pairs whose register changes, and is undefined if no pair
changes. The occupation component uses the no-pronoun condition. There,
$s_{\mathrm{occ}}$ is the population standard deviation of the 72
occupation-level honorific rates, each computed over the seven sentence frames
and measured in percentage points. For values bounded between 0 and 100, the
population standard deviation cannot exceed 50, so doubling it places both
components on a 0--100 scale, with higher values indicating more uneven
allocation of the respectful register. Because the gender component can be
unstable when few matched pairs change register, HBS should be read together
with HAI, GSR, and the significance tests, not as a significance test itself.

\subsection{Statistical analysis}
\label{sec:stats}

To compare gender and occupation on the same scale, we compute $\eta^2$
(eta-squared), which measures the proportion of variation in register choice
associated with each factor. For each model, we use matched masculine and
feminine items classified as honorific or ordinary, coding honorific as 1 and
ordinary as 0. We compute $\eta^2$ separately for occupation and pronoun.

We use McNemar's exact test \cite{mcnemar1947note} to determine whether the
directional gender difference is statistically significant. The masculine and
feminine conditions form matched pairs that differ only in the pronoun, making
McNemar's test appropriate. Because the test is applied to nine models, we use
Holm correction \cite{holm1979simple} to control the family-wise error rate at
$0.05$.

\textbf{Sensitivity analysis.}
We estimate the smallest directional gap detectable by the matched-pair design
using a standard normal approximation to McNemar's test. The estimate depends
on both the number of discordant pairs and the total number of usable pairs.
We use a nominal two-sided significance level of $0.05$ and 80\% power.
Because the reported significance tests are Holm-corrected across nine models,
these nominal thresholds should be interpreted as lower bounds on the detectable
effect size.

\begin{table}[t]
\caption{Gender and occupation effects across the nine models, sorted by
occupation spread. HAI and spread are reported in percentage points; GSR and
\textsc{mix} are percentages. $\eta^2$ gives the percentage of variance in
register choice associated with each factor. $p$ is the Holm-corrected
$p$-value from McNemar's exact test. HBS is the 0--100 Honorific Bias Score
defined in Section~\ref{sec:metrics}. \textbf{Bold} marks the minimum value
for GSR, spread, $\eta^2$, and HBS; for HAI, it marks the value closest to zero.}
\label{tab:main}

\centering
\footnotesize
\setlength{\tabcolsep}{2.5pt}

\begin{tabular}{lrrrr@{\hskip 5pt}rr@{\hskip 5pt}rr}
\toprule
& \multicolumn{4}{c@{\hskip 5pt}}{\textbf{Gender} (\textit{he}/\textit{she} pairs)}
& \multicolumn{2}{c@{\hskip 5pt}}{\textbf{Occupation}}
& \multicolumn{2}{c}{\textbf{Summary}} \\
\cmidrule(r{5pt}){2-5}
\cmidrule(r{5pt}){6-7}
\cmidrule{8-9}

Model & HAI & GSR & $p$ & $\eta^2$
      & spread & $\eta^2$ & HBS & \textsc{mix} \\
\midrule

Llama 4 Maverick
& $-7.19$ & 10.4 & $<.001$ & 0.50
& 35.5 & 51.5 & 70.1 & 0.5 \\

Llama 3.3 70B
& $+7.77$ & 11.6 & $<.001$ & 0.88
& 35.2 & 30.4 & 68.8 & 8.0 \\

Llama 4 Scout
& $+3.01$ & 9.8 & 0.269 & 0.11
& 33.7 & 29.2 & 49.0 & 4.1 \\

DeepSeek-V3.1
& $-0.84$ & 10.9 & 1.000 & 0.02
& 33.2 & 40.4 & 37.0 & 5.1 \\

Kimi K2 Thinking
& $-2.85$ & 12.5 & 0.556 & 0.10
& 30.7 & 25.0 & 42.1 & 6.2 \\

Gemini 2.5 Flash
& $+1.23$ & \textbf{3.7} & 0.952 & 0.01
& 28.5 & 29.3 & 45.2 & 4.9 \\

Qwen3-Next-80B
& $+3.32$ & 9.5 & 0.269 & 1.18
& 26.4 & 38.3 & 43.9 & 10.4 \\

gpt-oss-120b
& $\mathbf{+0.23}$ & 4.3 & 1.000 & \textbf{0.00}
& 17.3 & \textbf{3.2} & \textbf{19.9} & 8.6 \\

Gemma 4 26B
& $+1.20$ & 4.8 & 0.952 & 0.05
& \textbf{17.2} & 27.9 & 29.7 & 1.1 \\

\bottomrule
\end{tabular}
\end{table}

\section{Results}
\label{sec:results}

\textbf{Occupation effects.}
Occupation is associated with substantially greater variation in honorific use
than gender. For example, a model may translate ``The doctor will arrive at
nine'' using the honorific form \bn{আসবেন} (ashben), but ``The cleaner will
arrive at nine'' using the ordinary form \bn{আসবে} (ashbe), although neither
English sentence specifies a level of deference. We capture such differences
by computing the honorific rate for each occupation across the seven no-pronoun
sentence frames.

In the no-pronoun condition, the standard deviation of honorific rates across
the 72 occupations ranges from 17.2 percentage points for Gemma 4 26B to 35.5
for Llama 4 Maverick (Table~\ref{tab:main}). On the matched masculine and
feminine items, occupation is associated with 3.2\%--51.5\% of the variation
in register choice, compared with 0.00\%--1.18\% for the gender cue
($\eta^2$; Section~\ref{sec:stats}).

The nine models also rank occupations in much the same order. The mean pairwise
Spearman correlation between their occupation rankings is 0.77, ranging from
0.55 to 0.91. Leave-one-out checks show that the occupation spread does not
depend on any single occupation or sentence frame.

\textbf{Gender effects.}
Seven of the nine models show no significant directional gender difference
after Holm correction (Table~\ref{tab:main}). Five of them---Gemini 2.5 Flash,
Gemma 4 26B, gpt-oss-120b, DeepSeek-V3.1, and Kimi K2 Thinking---have HAI
values between $-2.85$ and $+1.23$ percentage points. Llama 4 Scout and
Qwen3-Next-80B show somewhat larger gaps of $+3.01$ and $+3.32$ percentage
points, but these are also not significant. Because the observed gaps fall
below the nominal detectable thresholds from the sensitivity analysis, we do
not treat non-significance as evidence that no gender effect exists.

Two models show significant effects in opposite directions. Llama 3.3 70B
favours men by 7.77 percentage points, whereas Llama 4 Maverick favours women
by 7.19 percentage points. Both remain significant after Holm correction.

HAI and GSR capture different aspects of gender sensitivity. Kimi K2 Thinking
and DeepSeek-V3.1, for example, change register relatively often, but these
changes do not consistently favour one gender. Their HAI values therefore stay
near zero despite comparatively high GSR values.

\textbf{Overall Honorific Bias Score.}
HBS ranges from 19.9 for gpt-oss-120b to 70.1 for Llama 4 Maverick. The two
models with significant directional gender effects, Llama 3.3 70B and
Llama 4 Maverick, also have the highest HBS values. Since HBS is a descriptive
summary rather than a significance test, we interpret it alongside HAI, GSR,
occupation spread, and $\eta^2$.

\textbf{Singular \textit{they}.}
Every model uses the respectful register less often for singular \textit{they}
than for the feminine condition. Most models translate \textit{they} using a
plural Bengali pronoun such as \bn{তারা} (\textit{tara}), with pluralisation
rates ranging from 45.8\% to 98.8\%. Among outputs that preserve singular
reference, the honorific rate increases in seven of the nine models and by more
than 25 percentage points in four. Llama 3.3 70B and Llama 4 Maverick are the
exceptions, as nearly all of their \textit{they} outputs become plural.

\section{Conclusion}

Bengali requires a translation system to choose a level of respect even when
the English source does not specify one. Using 2{,}016 sentences across 72
occupations, we evaluated nine models and found that honorific allocation varies
far more by occupation than by gender. In the no-pronoun condition, occupation
spread remains substantial across all models, and occupation is associated with
3.2\%--51.5\% of the variation in register choice, compared with
0.00\%--1.18\% for gender. The models also rank occupations in broadly similar
ways. Seven models show no significant directional gender difference after Holm
correction, while Llama 3.3 70B and Llama 4 Maverick show significant effects
in opposite directions. HBS ranges from 19.9 to 70.1, with these two models
also receiving the highest scores. Our findings are limited to controlled
template-based translation and do not establish how occupational patterns relate
to real-world social or gender distributions. Future work should test the method
on natural text and extend it to other South Asian languages that mark
honorificity.

\section*{Dataset Availability}
The evaluation set and the occupation list with its Bengali terms are publicly
available at
\href{https://github.com/ajwad-abrar/bangla-honorific-eval}{GitHub}.

\begin{credits}
\subsubsection{\discintname}
The authors have no competing interests to declare that are relevant to the
content of this article.
\end{credits}

\bibliographystyle{splncs04}
\bibliography{references}

\end{document}